\pdfoutput=1

\documentclass[11pt]{article}

\usepackage[]{acl}

\usepackage{times}
\usepackage{latexsym}
\usepackage{expex}
\usepackage{subcaption}
\usepackage{graphicx}
\usepackage[T1]{fontenc}

\usepackage[utf8]{inputenc}

\usepackage{microtype}

\usepackage{tabularx}
\usepackage{float}

\title{Incorporating Annotator Uncertainty into Representations of Discourse Relations}

\author{S. Magalí López Cortez \qquad 
Cassandra L. Jacobs \\
  Department of Linguistics \\
  University at Buffalo \\
  Buffalo, NY, USA \\
  \texttt{solmagal;cxjacobs@buffalo.edu} \\}

\begin{document}
\maketitle
\begin{abstract}
Annotation of discourse relations is a known difficult task, especially for non-expert annotators. 
In this paper, we investigate novice annotators' uncertainty on the annotation of discourse relations on spoken conversational data. 
We find that dialogue context (single turn, pair of turns within speaker, and pair of turns across speakers) is a significant predictor of  confidence scores.
We compute distributed representations of discourse relations from co-occurrence statistics that incorporate information about confidence scores and dialogue context. 
We perform a hierarchical clustering analysis using these representations and show that weighting discourse relation representations with information about confidence and dialogue context coherently models our annotators' uncertainty about discourse relation labels.
\end{abstract}

\section{Introduction}

Discourse relations (DRs) are those relations such as Elaboration, Explanation, Narration, which hold between discourse units. 
The task of labeling DRs is known to pose difficulties for annotators \citep{spooren2010coding}, as sometimes more than one interpretation may be possible \citep{scholman-etal-2022-discogem,webber-2013-excludes}. 

Recent studies have shown that allowing for multiple labels in annotation can improve the performance of discourse parsers \citep{yung-etal-2022-label}. 
\citet{scholman-etal-2022-discogem} test different label aggregation methods in a crowdsourced corpus annotated by 10 workers and find that probability distributions over labels better capture ambiguous interpretations of discourse relations than majority class labels. 
(\ref{discogemeg}) shows an example from their corpus, where the relation between the second and third sentences (in italics and bold, respectively), was interpreted as Conjunction by four annotators  and Result by five annotators. 

\ex \label{discogemeg} It is logical that our attention is focused on cities. \textit{Cities are home to 80\% of the 500 million or so inhabitants of the EU.}  \textbf{It is in cities that the great majority of jobs, companies and centres of education are located.} \citep[adapted from DiscoGeM, Europarl genre;][italics and bolding are ours.]{scholman-etal-2022-discogem} 
\xe

Annotating the discourse relation between these two sentences with both Conjunction and Result captures different possible interpretations of the relation between these segments. 
For example, the two sentences may contain two conjoined facts about cities, but can also be perceived as describing a causal relation between the first and second sentence (i.e., as cities are home to the largest part of the population, most jobs, companies and educational institutions are located there).

In this work, we investigate which relations are distributionally similar or co-occurring in multilabel annotations of spontaneous conversations. 
We are particularly interested in how novice annotators interpret discourse relation categories when annotating spoken conversational data. 
We collect annotations of DRs from Switchboard telephone conversations \citep{10.5555/1895550.1895693}, allowing for multiple labels, and ask for confidence scores. 
We find that confidence scores vary significantly across dialogue contexts (single turn vs. pairs of turns produced by the same speaker vs. pairs of turns produced by different speakers). 
We incorporate information about these three dialogue context types and confidence scores into distributed representations of discourse relations. 
A clustering analysis shows that discourse relations that tend to occur across speakers cluster together, while discourse relations which tend to occur within a speaker, either in the same turn or different turns, form their own cluster.

\section{Annotation of Discourse Relations}
Our analyses are built on the dataset collected in \citet{lopez-cortez-jacobs-2023-distribution}, who selected 19 conversations from  Switchboard\footnote{We discarded the annotations from one conversation because the annotators did not follow the guidelines.}, a corpus consisting of telephone conversations between pairs of participants about a variety of topics (e.g. recycling, movies, child care). 
We chose this corpus because it contains informal, spontaneous dialogues, and because it has been used within linguistics in various studies on conversation \citep{jaeger2013alignment,reitter2014alignment}.

\subsection{Discourse Units}
An initial set of turns for annotation was selected by using spaCy’s dependency parser \cite[][version 3.3.1]{spacy} to select turns with two or more \textsc{ROOT} or \textsc{VERB} tags. 
We define a turn as each segment of dialogue taken from Switchboard. 
We note that an utterance produced by one speaker (A) may take place during a continuous utterance by another speaker (B).
Switchboard splits A's utterance into two turns in these cases. 
We return to this point in the Discussion. 

We manually segmented these turns into elementary discourse units (EDUs). 
The main criteria for segmenting turns into EDUs was that the unit performs some basic discourse function \citep{asher2003logics}. 
By default, finite clauses are considered EDUs, as well as comment words like “Really?” or acknowledgments such as “Uh-huh” or “Yeah.”  
Cases of interruptions and repairs were segmented if they constituted a turn in Switchboard, as in example (\ref{interruption}), and when they contained a verb, as in example (\ref{verb}). 
Cases of repetition as in (\ref{repetition}) were not considered separate EDUs. 
We segmented disfluencies (“uh”) and some non-verbal communication (“[laughter]”) but we did not select these for discourse relation labeling.

\pex 
\a \label{interruption} B: || So you don't see too many thrown out around the || [laughter] || streets. ||\\
A: || Really || \\
B: || Or even bottles. ||

\a \label{verb}
 B: || I think, || uh, || I wonder || if that worked. ||

\a \label{repetition}
A: || What kind of experience do you, do you have, then with child care? || 
\xe

Because many EDUs are very short, we selected pairs of elementary discourse units and complex discourse units (CDUs) for discourse relation annotation. 
CDUs consist of two or more EDUs that constitute an argument to a discourse relation \cite{asher2003logics}. We use the term \textit{discourse units} (DUs) to refer to both EDUs and CDUs.

\subsection{Dialogue Contexts}

We manually selected items for annotation across three different contexts: within a single turn, across two turns within a speaker, and across two immediately adjacent turns (two speakers). 
(\ref{EDUsKinds}) shows an example for each context kind, with the first DU in italics and the second in bold. 
Example (\ref{within}) shows two discourse units within a speaker’s turn. (\ref{across-same}) shows two discourse units uttered by the same speaker but that span across two different turns, interrupted by one turn. 
We did not include any constraint for the length of the interrupting turn. 
(\ref{across-dif}) shows two DUs uttered by speakers in adjacent turns. 
We leave for future work the annotation of pairs of discourse units that may have a longer-distance relation with more turns in between DUs.

\pex \label{EDUsKinds}
\a \label{within} A: || \textit{and they discontinued them} || \textbf{because people were coming and dumping their trash in them.} || 
\a \label{across-same} 
B: || No, || \textit{I just, I noticed || in Iowa and other cities like that, it's a nickel per aluminum can.} ||\\
A: || Oh. ||\\
B: || \textbf{So you don't see too many thrown out around the || [laughter] || streets}. 
\a \label{across-dif} 
A: || \textit{We live in the Saginaw area.} || \\ B: || \textbf{Saginaw?} || 
\xe

\subsection{Taxonomy of Discourse Relations}
The DRs chosen to annotate our corpus were adapted from the STAC corpus manual \citep{asher2012manual,asher-etal-2016-discourse}. STAC is a corpus of strategic multi-party chat conversations in an online game. 
Table \ref{table:relstaxonomy} shows the taxonomy used. 
We selected 11 DRs based on a pilot annotation by the first author, and added an “Other” category for relations not included in the list of labels. 
We focused on a small taxonomy to minimize the number of choices presented to our novice annotators. 
We refer readers to \citet{lopez-cortez-jacobs-2023-distribution} for details and examples of each relation in the taxonomy. 
Future work will include revising the taxonomy used. 

\begin{table}[h]
\begin{center}
\begin{tabular}{l|l}
    \hline
 Acknowledgement & Elaboration\\ 
 Background & Explanation\\ 
 Clarification Question & Narration\\ 
 Comment & Question-Answer Pair \\ 
 Continuation & Result\\ 
 Contrast & Other\\ 
\hline

\end{tabular}
\end{center}

\caption{Taxonomy of discourse relations.}
\label{table:relstaxonomy}
\end{table}

\subsection{Annotation Procedure}

The annotation of discourse relations was done by students enrolled in a Computational Linguistics class. 
Students were divided into 19 teams of approximately 5 members each, and each team was assigned a conversation. 
The annotation was performed individually, but teams then discussed their work and wrote a report together. 
Annotators were trained using written guidelines, a quiz-like game, and a live group annotation demo. 

We used the annotation interface Prodigy \cite{montani2018prodigy}. 
Each display presented the two target discourse units plus two context turns before and two after. 
Annotators also had access to the entire conversation throughout the annotation task.
Below the text, the screen showed a multiple choice list of discourse relations plus the “Other” category. 
We allowed for the selection of multiple labels following previous findings that allowing for multiple labels better captures ambiguous interpretations of discourse relations \cite{scholman-etal-2022-discogem} and improves the performance of discourse parsers \cite{yung-etal-2022-label}.

Each display also asked for confidence scores in the range 1-5, corresponding to least to most confident. 
We did not pursue label-specific confidence scores but rather the confidence in the label(s) as a whole in the interest of minimizing annotator overhead. 
The results of this work show that per-label confidence scores or a slider-based approach may be informative and is a topic for future work. We include an example annotation item in Appendix \ref{interface}.
 
\section{Dialogue Context as a Predictor of Confidence Scores}

First we sought to understand how discourse relations and dialogue context (as defined above) influence annotator confidence. 
Because our confidence ratings data has multiple observations for each annotator, each team and each DU, it is hierarchical and thus benefits from being analyzed using hierarchical mixed effects models. 
Due to the ordinal nature of the ratings data, we use the cumulative link approach (CLMM; \citealp{liddell2018analyzing,howcroft-rieser-2021-happens}) rather than model confidence scores as real-valued in linear regression.
We first built a null model containing only random intercepts by annotator and compared it to a model containing an additional fixed effect and random slope by annotator for dialogue context type: single turn, across turns within speaker and across speakers ($kind$, dummy coded).
A likelihood ratio test revealed a significant improvement in fit by adding kind as a predictor ($\chi^2(7) = 126.64, p<0.001$). 
Adding random intercepts for DU pairs to account for annotation difficulty across DU pairs also led to a significant improvement in model fit beyond the model containing dialogue context kind ($\chi^2(1)= 195.01, p<0.001$). 
This suggests that our annotators' confidence scores are sensitive to the context of DU pairs.

\begin{figure}[h]
    \centering   \includegraphics[width=.9\columnwidth]{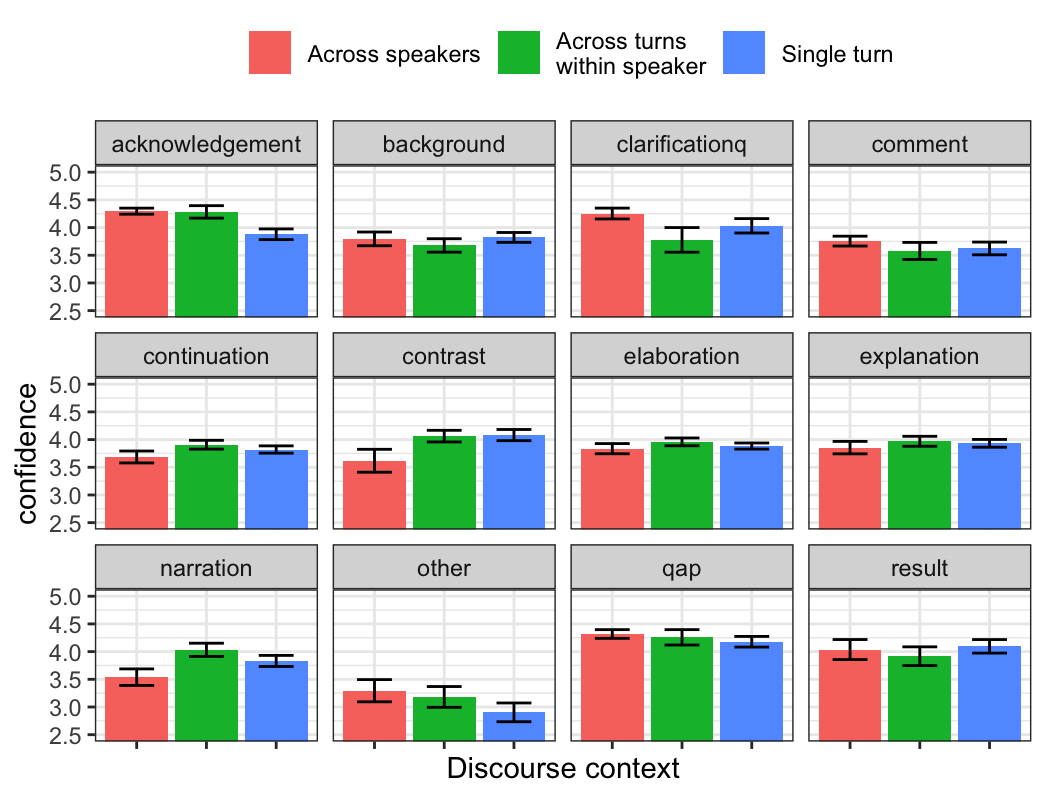}
    \caption{Confidence scores per context kind across discourse relations. \textit{qap} stands for Question-Answer Pair and \textit{clarificationq} for Clarification Question.}
    \label{fig:confidence}
\end{figure}

Figure \ref{fig:confidence} shows mean confidence scores per context kind across discourse relations. 
Confidence scores within a speaker both across and within turns received similar confidence ratings ($\beta = -0.13$, $z = -0.56$, $p =$ n.s.\footnote{Not statistically significant.}), while annotators were significantly more confident for relation annotation across speakers ($\beta = 0.63$, $z = 3.05$, $p < .01$).
The CLMM revealed that annotators used confidence scores between 3 and 5 overall, except for the label “Other”, for which they selected lower confidence scores.
Background received lower confidence scores overall. Continuation, Contrast and Narration received higher scores for contexts within speaker. Comment and Result received higher scores for turns across speakers and single turn. 
For Elaboration and Explanation, mean confidence scores are very similar across the three contexts, with slightly higher scores for single turn and pairs of turns within speaker. 
Acknowledgment, Clarification Question (“clarificationq”) and Question-Answer Pair (“qap”) received higher scores for turns across speakers, which makes sense given the dialogic nature of these relations. 
However, these relations also received rather high confidence scores for single turn and pairs of turns within speaker, which is a bit surprising. 
We suspect this might be due to the context turns included for each pair of DUs, which might have led annotators to choose relations between discourse units other than for the pair of highlighted DUs. 
Future analysis will look closer at this aspect.

\section{Distributed Representations from Discourse Relation Annotations}

To model the similarity between discourse relations as perceived by annotators, we computed embedding representations of discourse relations. 
We extracted each $n$ individual annotation containing \textbf{r}elation-\textbf{c}onfidence $(r, c)$ tuples selected by a given annotator for a pair of DUs.
We concatenate bag-of-relation vectors with one-hot encoded features representing the dialogue context kind, and multiply the count vector of annotated relations (either 1 or 0 for each relation) by the confidence score (1-5) for that pair of DUs.
This weighting learns more from high confidence; an ideal reweighting may be possible with additional parameter search, possibly in conjunction with the CLMM outputs.

For an $n \times 1$ confidence ratings matrix $C$, an $n \times 12$ bag-of-relations matrix $R$, an $n \times 3$ discourse context matrix $D$ for each annotation, we obtain an annotation matrix $A = C \times (R | D)$.
We then obtain a square co-occurrence matrix $O$ such that $O = A \cdot A^{T}$, which we factorize using Principal Component Analysis (without shifting the intercept following \citealp{levy2014neural}).
Each relation is thus represented as a vector that consolidates co-occurrences between all relations within a single annotator that are weighted by confidence score. 
We then projected these embeddings into two dimensions with UMAP \cite{mcinnes2018umap} and performed a hierarchical clustering analysis over the resulting coordinates due to the greater discriminability afforded by continuous distance metrics.

\begin{figure*}[h]
    \begin{subfigure}[t]{.475\textwidth}
    \centering
    \includegraphics[width=\linewidth]{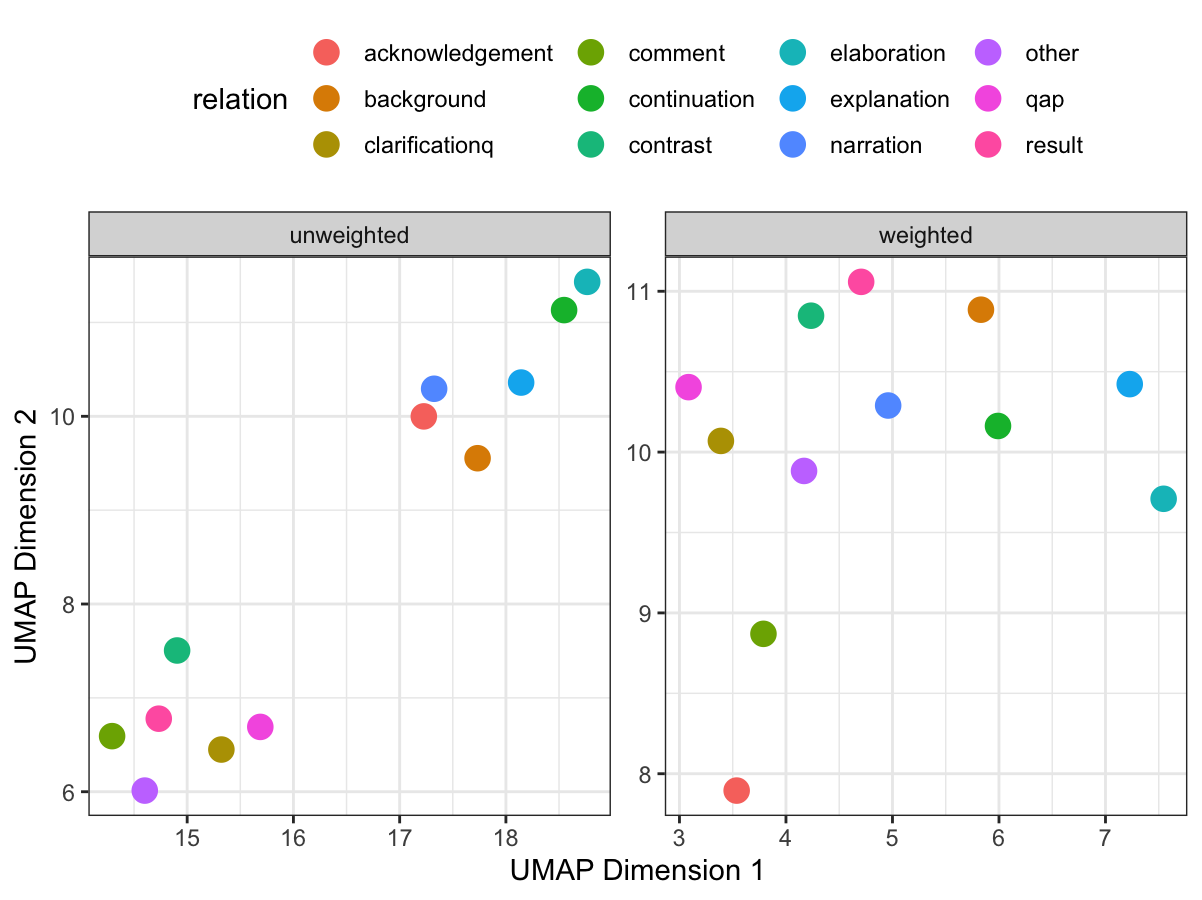}
    \caption[width=.8\linewidth]{The coordinates obtained with UMAP for all discourse relations plotted in two-dimensional space. The plot on the left shows the unweighted embedding representations and the figure on the right shows the weighted embedding representations.}
    \label{fig:umap}
    \end{subfigure}
    \hfill
    \begin{subfigure}[t]{.475\textwidth}
    \centering
    \includegraphics[width=1  \linewidth]{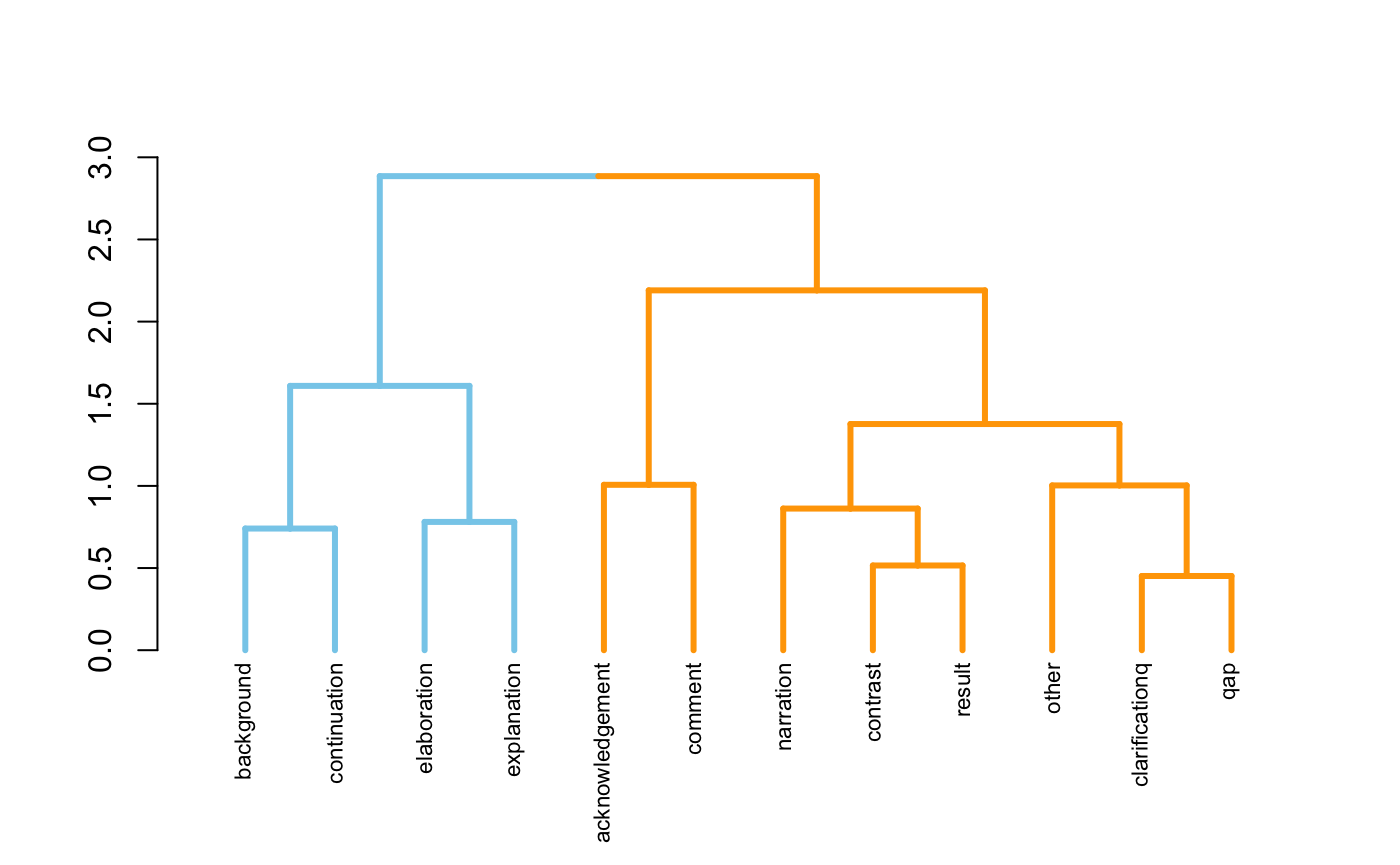}
    \caption[width=.8\linewidth]{Dendrogram showing hierarchical clustering of Discourse Relations built from UMAP coordinates. \textit{qap} stands for Question-Answer Pair and \textit{clarificationq} for Clarification Question.}
    \label{fig:dendrogram}
    \end{subfigure}
    \caption{Dimensionality reduction and clustering of relation embeddings.}
\end{figure*}

Informally, the UMAP coordinates appear more gradient in the representational space when confidence was included (right panel) than when it was not included (left panel).
When context is not included, the UMAP coordinates primarily represent the frequency of labels in our corpus, which we include in Appendix \ref{appendix-frequencies}.
We visualize the UMAP coordinates in Figure \ref{fig:umap}.
Figure \ref{fig:dendrogram} shows a dendrogram with the output clusters, colored according to the optimal number of clusters ($k=2$), calculated using average silhouette widths \cite{levshina2022semantic}. 
There are two large clusters, one of which contains two sub-clusters with Background and Continuation, on the one hand, and Elaboration and Explanation on the other. 
In the other large cluster, Acknowledgement and Comment form a sub-cluster.
These are very common relations between pairs of turns across speakers. 
Clarification Question and Question-Answer Pair form another sub-cluster, also common relations between pairs of turns across speakers, in close proximity to the Other label, which received a sub-cluster of its own. 
Narration and Contrast and Result, form the last sub-clusters, which we suspect is due in part to the frequencies of these relations \cite{schnabel-etal-2015-evaluation}. We include a dendrogram with the output clusters of a hierarchical clustering analysis performed with base bag-of-relations vectors (without context kind and confidence scores weight) in Figure \ref{fig:clust_no_conf} in Appendix \ref{clustering-2} for comparison.

Currently, we provide these results as a proof of concept of the feasibility and interpretability of noisy labels produced by novice annotators.
Importantly, annotations weighted by confidence produce coherent clusters of discourse relations.
We envision applications of DR embeddings to several domains including dialogue generation, such that appropriate responses to input are partially conditioned on a latent or mixed combination of DRs.

\section{Related Work}

Annotation of discourse relations is usually done within Rhetorical Structure Theory \citep{mann1987rhetorical}, as in the RST-DT \citep{carlson2003building} and GUM \citep{zeldes2017gum} corpora, within Segmented Discourse Representation Theory \citep[SDRT,][]{asher2003logics}, as in the STAC \citep{asher-etal-2016-discourse} and Molweni \citep{li-etal-2020-molweni} corpora, or within the Penn Discourse Treebank framework \citep{prasad-etal-2008-penn,prasad2014reflections,prasad-etal-2018-discourse}. We use a taxonomy adapted from SDRT, in particular, the STAC corpus.

Annotators are usually trained to identify discourse relations using the framework's taxonomy. 
Some recent alternatives to explicitly collecting annotation of DRs include crowdsourcing by eliciting connectives \citep{yung-etal-2019-crowdsourcing,scholman-etal-2022-discogem} or question-answer pairs \citep{pyatkin-etal-2020-qadiscourse} rather than relations. 
In this work, we wanted to investigate how annotators perceive discourse relation categories, and therefore a connective insertion task would only provide indirect evidence. 
We train annotators on DR labeling and ask annotators to choose from a set of discourse relation labels. 
We allow for multiple labels to investigate what relations are more confusable or perceived as co-occurring \citep{marchal-etal-2022-establishing}.

\section{Discussion and Future Work}
In this study, we collected multiple annotations of discourse relations from a subset of the Switchboard corpus, together with confidence scores. 
We found that dialogue context had a significant effect on confidence scores. 
We computed embedding representations of DRs using co-occurrence statistics and weighted the vectors using context type and confidence scores, and found that these representations coherently model our annotators uncertainty about discourse relation labels.

Discourse units that occur across turns as defined by Switchboard do not necessarily occur across continuous utterances from the speaker's point-of-view. 
Obtaining information about whether same-speaker pairs of discourse units fall into the same or different utterances may help to explain additional variance in annotator confidence.

Additionally, in this work, we investigated annotators' confidence on the annotation of adjacent turns. 
In future work, we plan to annotate discourse relations across longer-distance discourse units and to allow for hierarchical annotation. 
We expect that annotation confidence will also vary across longer-distance units and across different depths of annotation.

In the future, we plan to use this information to run a larger scale annotation study of the Switchboard corpus to analyze discourse relation patterns in spoken dialogues.

\section*{Limitations}
This work is limited by the size of the dataset and the taxonomy used in the annotation task. 
While we found that our annotators perceived some of the categories as more similar or confusable, future work can examine annotators’ uncertainty in a larger set of discourse relations. 
The selection of DUs for annotation was also non-exhaustive. 
In future work, we plan to expand the selection procedure so that we include more distantly related DUs. 
We also note that the frequency of discourse relation labels and individual differences in confidence levels among annotators may bias the representations. We plan to look into these potential biases in future work.

\section*{Ethics Statement}
We are not aware of ethical issues associated with the texts used in this work. 
Students participated in the annotation task as part of course credit but annotation decisions were not associated with their performance in the course.

\section*{Acknowledgements}
We would like to thank Jürgen Bohnemeyer and three anonymous reviewers for feedback on a previous version of this paper. We also thank the students who participated in the annotation task.

\bibliography{anthology,custom}

\begin{thebibliography}{29}
\expandafter\ifx\csname natexlab\endcsname\relax\def\natexlab#1{#1}\fi

\bibitem[{Asher et~al.(2016)Asher, Hunter, Morey, Farah, and
  Afantenos}]{asher-etal-2016-discourse}
Nicholas Asher, Julie Hunter, Mathieu Morey, Benamara Farah, and Stergos
  Afantenos. 2016.
\newblock \href {https://aclanthology.org/L16-1432} {Discourse structure and
  dialogue acts in multiparty dialogue: the {STAC} corpus}.
\newblock In \emph{Proceedings of the Tenth International Conference on
  Language Resources and Evaluation ({LREC}'16)}, pages 2721--2727,
  Portoro{\v{z}}, Slovenia. European Language Resources Association (ELRA).

\bibitem[{Asher and Lascarides(2003)}]{asher2003logics}
Nicholas Asher and Alex Lascarides. 2003.
\newblock \emph{Logics of conversation}.
\newblock Cambridge University Press.

\bibitem[{Asher et~al.(2012)Asher, Popescu, Muller, Afantenos, Cadilhac,
  Benamara, Vieu, and Denis}]{asher2012manual}
Nicholas Asher, Vladimir Popescu, Philippe Muller, Stergos Afantenos, Anais
  Cadilhac, Farah Benamara, Laure Vieu, and Pascal Denis. 2012.
\newblock Manual for the analysis of settlers data.
\newblock \emph{Strategic Conversation (STAC). Universit{\'e} Paul Sabatier}.

\bibitem[{Carlson et~al.(2003)Carlson, Marcu, and
  Okurowski}]{carlson2003building}
Lynn Carlson, Daniel Marcu, and Mary~Ellen Okurowski. 2003.
\newblock Building a discourse-tagged corpus in the framework of rhetorical
  structure theory.
\newblock In \emph{Current and new directions in discourse and dialogue}, pages
  85--112. Springer.

\bibitem[{Godfrey et~al.(1992)Godfrey, Holliman, and
  McDaniel}]{10.5555/1895550.1895693}
John~J. Godfrey, Edward~C. Holliman, and Jane McDaniel. 1992.
\newblock {SWITCHBOARD: Telephone Speech Corpus for Research and Development}.
\newblock In \emph{Proceedings of the 1992 IEEE International Conference on
  Acoustics, Speech and Signal Processing - Volume 1}, ICASSP'92, page
  517–520, USA. IEEE Computer Society.

\bibitem[{Honnibal et~al.(2020)Honnibal, Montani, Van~Landeghem, and
  Boyd}]{spacy}
Matthew Honnibal, Ines Montani, Sofie Van~Landeghem, and Adriane Boyd. 2020.
\newblock \href {https://doi.org/10.5281/zenodo.8123552} {{"spaCy:
  Industrial-strength Natural Language Processing in Python"}}.

\bibitem[{Howcroft and Rieser(2021)}]{howcroft-rieser-2021-happens}
David~M. Howcroft and Verena Rieser. 2021.
\newblock \href {https://doi.org/10.18653/v1/2021.emnlp-main.703} {What happens
  if you treat ordinal ratings as interval data? human evaluations in {NLP} are
  even more under-powered than you think}.
\newblock In \emph{Proceedings of the 2021 Conference on Empirical Methods in
  Natural Language Processing}, pages 8932--8939, Online and Punta Cana,
  Dominican Republic. Association for Computational Linguistics.

\bibitem[{Jaeger and Snider(2013)}]{jaeger2013alignment}
T~Florian Jaeger and Neal~E Snider. 2013.
\newblock Alignment as a consequence of expectation adaptation: Syntactic
  priming is affected by the prime’s prediction error given both prior and
  recent experience.
\newblock \emph{Cognition}, 127(1):57--83.

\bibitem[{Levshina(2022)}]{levshina2022semantic}
Natalia Levshina. 2022.
\newblock Semantic maps of causation: New hybrid approaches based on corpora
  and grammar descriptions.
\newblock \emph{Zeitschrift f{\"u}r Sprachwissenschaft}, 41(1):179--205.

\bibitem[{Levy and Goldberg(2014)}]{levy2014neural}
Omer Levy and Yoav Goldberg. 2014.
\newblock \href
  {https://proceedings.neurips.cc/paper_files/paper/2014/file/feab05aa91085b7a8012516bc3533958-Paper.pdf}
  {Neural word embedding as implicit matrix factorization}.
\newblock \emph{Advances in neural information processing systems}, 27.

\bibitem[{Li et~al.(2020)Li, Liu, Kan, Zheng, Wang, Lei, Liu, and
  Qin}]{li-etal-2020-molweni}
Jiaqi Li, Ming Liu, Min-Yen Kan, Zihao Zheng, Zekun Wang, Wenqiang Lei, Ting
  Liu, and Bing Qin. 2020.
\newblock \href {https://doi.org/10.18653/v1/2020.coling-main.238} {Molweni: A
  challenge multiparty dialogues-based machine reading comprehension dataset
  with discourse structure}.
\newblock In \emph{Proceedings of the 28th International Conference on
  Computational Linguistics}, pages 2642--2652, Barcelona, Spain (Online).
  International Committee on Computational Linguistics.

\bibitem[{Liddell and Kruschke(2018)}]{liddell2018analyzing}
Torrin~M Liddell and John~K Kruschke. 2018.
\newblock Analyzing ordinal data with metric models: What could possibly go
  wrong?
\newblock \emph{Journal of Experimental Social Psychology}, 79:328--348.

\bibitem[{L{\'o}pez~Cortez and
  Jacobs(2023)}]{lopez-cortez-jacobs-2023-distribution}
S.~Magal{\'\i} L{\'o}pez~Cortez and Cassandra~L. Jacobs. 2023.
\newblock \href {https://aclanthology.org/2023.codi-1.21} {The distribution of
  discourse relations within and across turns in spontaneous conversation}.
\newblock In \emph{Proceedings of the 4th Workshop on Computational Approaches
  to Discourse (CODI 2023)}, pages 156--162, Toronto, Canada. Association for
  Computational Linguistics.

\bibitem[{Mann and Thompson(1987)}]{mann1987rhetorical}
William~C Mann and Sandra~A Thompson. 1987.
\newblock \emph{Rhetorical structure theory: A theory of text organization}.
\newblock University of Southern California, Information Sciences Institute Los
  Angeles.

\bibitem[{Marchal et~al.(2022)Marchal, Scholman, Yung, and
  Demberg}]{marchal-etal-2022-establishing}
Marian Marchal, Merel Scholman, Frances Yung, and Vera Demberg. 2022.
\newblock \href {https://aclanthology.org/2022.coling-1.322} {Establishing
  annotation quality in multi-label annotations}.
\newblock In \emph{Proceedings of the 29th International Conference on
  Computational Linguistics}, pages 3659--3668, Gyeongju, Republic of Korea.
  International Committee on Computational Linguistics.

\bibitem[{McInnes et~al.(2018)McInnes, Healy, Saul, and
  Gro{\ss}berger}]{mcinnes2018umap}
Leland McInnes, John Healy, Nathaniel Saul, and Lukas Gro{\ss}berger. 2018.
\newblock {UMAP}: Uniform manifold approximation and projection.
\newblock \emph{Journal of Open Source Software}, 3(29):861.

\bibitem[{Montani and Honnibal(2018)}]{montani2018prodigy}
Ines Montani and Matthew Honnibal. 2018.
\newblock Prodigy: A new annotation tool for radically efficient machine
  teaching.
\newblock \emph{Artificial Intelligence}.

\bibitem[{Prasad et~al.(2008)Prasad, Dinesh, Lee, Miltsakaki, Robaldo, Joshi,
  and Webber}]{prasad-etal-2008-penn}
Rashmi Prasad, Nikhil Dinesh, Alan Lee, Eleni Miltsakaki, Livio Robaldo,
  Aravind Joshi, and Bonnie Webber. 2008.
\newblock \href
  {http://www.lrec-conf.org/proceedings/lrec2008/pdf/754_paper.pdf} {The {P}enn
  {D}iscourse {T}ree{B}ank 2.0.}
\newblock In \emph{Proceedings of the Sixth International Conference on
  Language Resources and Evaluation ({LREC}'08)}, Marrakech, Morocco. European
  Language Resources Association (ELRA).

\bibitem[{Prasad et~al.(2014)Prasad, Webber, and Joshi}]{prasad2014reflections}
Rashmi Prasad, Bonnie Webber, and Aravind Joshi. 2014.
\newblock Reflections on the {P}enn {D}iscourse {T}reebank, {C}omparable
  {C}orpora, and {C}omplementary {A}nnotation.
\newblock \emph{Computational Linguistics}, 40(4):921--950.

\bibitem[{Prasad et~al.(2018)Prasad, Webber, and
  Lee}]{prasad-etal-2018-discourse}
Rashmi Prasad, Bonnie Webber, and Alan Lee. 2018.
\newblock \href {https://aclanthology.org/W18-4710} {Discourse annotation in
  the {PDTB}: The next generation}.
\newblock In \emph{Proceedings 14th Joint {ACL} - {ISO} Workshop on
  Interoperable Semantic Annotation}, pages 87--97, Santa Fe, New Mexico, USA.
  Association for Computational Linguistics.

\bibitem[{Pyatkin et~al.(2020)Pyatkin, Klein, Tsarfaty, and
  Dagan}]{pyatkin-etal-2020-qadiscourse}
Valentina Pyatkin, Ayal Klein, Reut Tsarfaty, and Ido Dagan. 2020.
\newblock \href {https://doi.org/10.18653/v1/2020.emnlp-main.224}
  {{QAD}iscourse - {D}iscourse {R}elations as {QA} {P}airs: {R}epresentation,
  {C}rowdsourcing and {B}aselines}.
\newblock In \emph{Proceedings of the 2020 Conference on Empirical Methods in
  Natural Language Processing (EMNLP)}, pages 2804--2819, Online. Association
  for Computational Linguistics.

\bibitem[{Reitter and Moore(2014)}]{reitter2014alignment}
David Reitter and Johanna~D Moore. 2014.
\newblock Alignment and task success in spoken dialogue.
\newblock \emph{Journal of Memory and Language}, 76:29--46.

\bibitem[{Schnabel et~al.(2015)Schnabel, Labutov, Mimno, and
  Joachims}]{schnabel-etal-2015-evaluation}
Tobias Schnabel, Igor Labutov, David Mimno, and Thorsten Joachims. 2015.
\newblock \href {https://doi.org/10.18653/v1/D15-1036} {Evaluation methods for
  unsupervised word embeddings}.
\newblock In \emph{Proceedings of the 2015 Conference on Empirical Methods in
  Natural Language Processing}, pages 298--307, Lisbon, Portugal. Association
  for Computational Linguistics.

\bibitem[{Scholman et~al.(2022)Scholman, Dong, Yung, and
  Demberg}]{scholman-etal-2022-discogem}
Merel Scholman, Tianai Dong, Frances Yung, and Vera Demberg. 2022.
\newblock \href {https://aclanthology.org/2022.lrec-1.351} {{D}isco{G}e{M}: A
  crowdsourced corpus of genre-mixed implicit discourse relations}.
\newblock In \emph{Proceedings of the Thirteenth Language Resources and
  Evaluation Conference}, pages 3281--3290, Marseille, France. European
  Language Resources Association.

\bibitem[{Spooren and Degand(2010)}]{spooren2010coding}
Wilbert Spooren and Liesbeth Degand. 2010.
\newblock Coding coherence relations: Reliability and validity.
\newblock \emph{Corpus Linguistics and Linguistic Theory}, 6(2):241--266.

\bibitem[{Webber(2013)}]{webber-2013-excludes}
Bonnie Webber. 2013.
\newblock \href {https://aclanthology.org/W13-0124} {What excludes an
  alternative in coherence relations?}
\newblock In \emph{Proceedings of the 10th International Conference on
  Computational Semantics ({IWCS} 2013) {--} Long Papers}, pages 276--287,
  Potsdam, Germany. Association for Computational Linguistics.

\bibitem[{Yung et~al.(2022)Yung, Anuranjana, Scholman, and
  Demberg}]{yung-etal-2022-label}
Frances Yung, Kaveri Anuranjana, Merel Scholman, and Vera Demberg. 2022.
\newblock \href {https://aclanthology.org/2022.codi-1.7} {Label distributions
  help implicit discourse relation classification}.
\newblock In \emph{Proceedings of the 3rd Workshop on Computational Approaches
  to Discourse}, pages 48--53, Gyeongju, Republic of Korea and Online.
  International Conference on Computational Linguistics.

\bibitem[{Yung et~al.(2019)Yung, Demberg, and
  Scholman}]{yung-etal-2019-crowdsourcing}
Frances Yung, Vera Demberg, and Merel Scholman. 2019.
\newblock \href {https://doi.org/10.18653/v1/W19-4003} {Crowdsourcing discourse
  relation annotations by a two-step connective insertion task}.
\newblock In \emph{Proceedings of the 13th Linguistic Annotation Workshop},
  pages 16--25, Florence, Italy. Association for Computational Linguistics.

\bibitem[{Zeldes(2017)}]{zeldes2017gum}
Amir Zeldes. 2017.
\newblock {The GUM corpus: Creating multilayer resources in the classroom}.
\newblock \emph{Language Resources and Evaluation}, 51(3):581--612.

\end{thebibliography}
\bibliographystyle{acl_natbib}

\appendix
\section{Frequencies of Discourse Relation Labels} \label{appendix-frequencies}

\begin{table}[H]
    \centering
    \begin{tabular}{lr}
    \textbf{Discourse Relation} & \textbf{Count} \\
    \hline
        Elaboration & 636 \\
        Continuation & 554 \\
        Acknowledgement & 494 \\
        Explanation & 383 \\
        Comment & 265 \\
        Background & 252 \\
        Narration & 249 \\
        Question-Answer Pair & 248 \\
        Contrast & 191 \\
        Clarification Question & 179 \\
        Result & 124 \\
        Other & 106 \\ 
        \hline
    \end{tabular}
    \caption{Raw counts of discourse relation labels in our corpus from most to least frequent.}
    \label{tab:rel-frequencies}
\end{table}

\section{Clustering without Context and Confidence Weighting} \label{clustering-2}

\begin{figure}[H]
    \centering   \includegraphics[width=1\columnwidth]{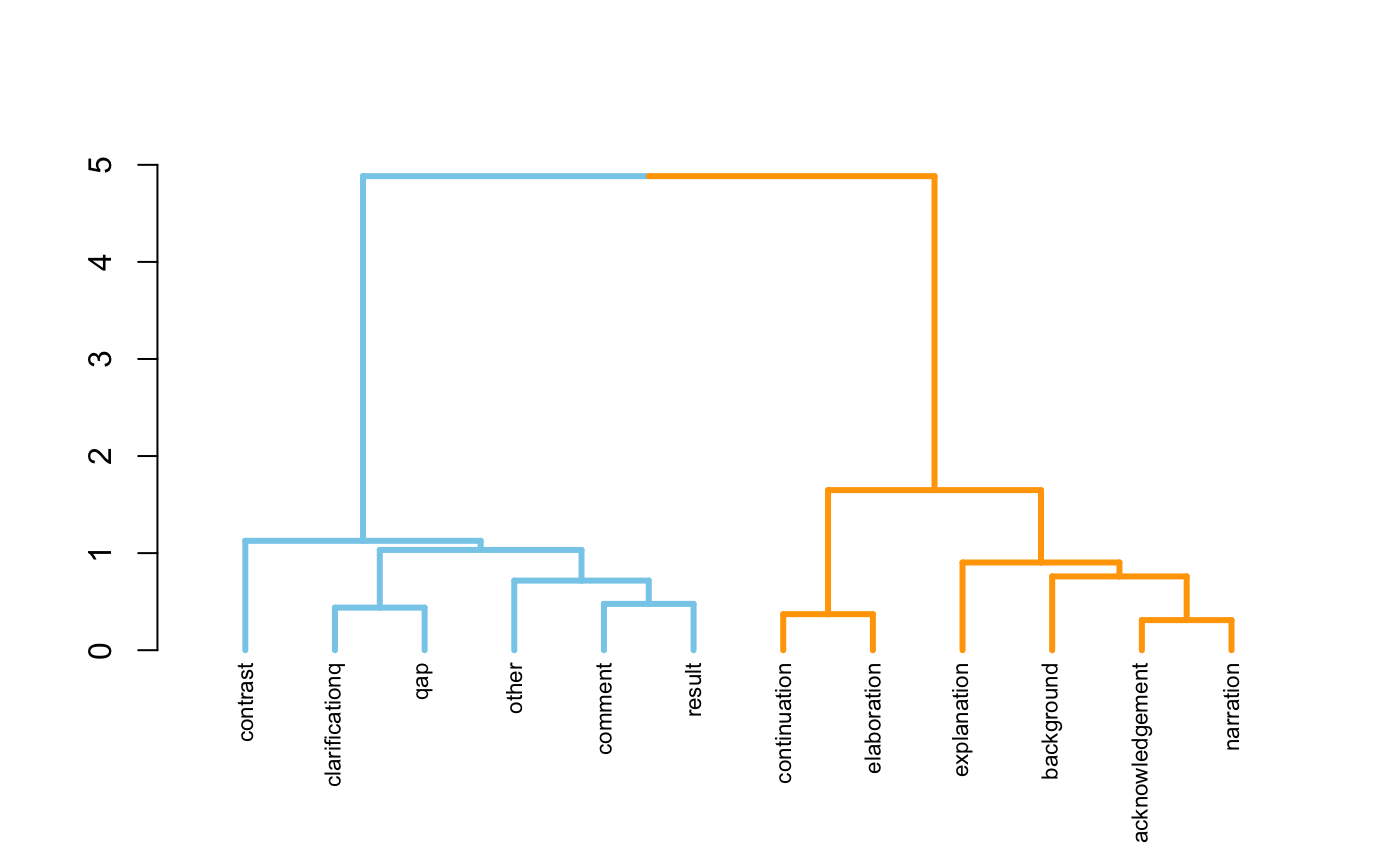}
    \caption{Dendrogram showing hierarchical clustering of Discourse Relations built from UMAP coordinates without context kind and confidence scores weighting. \textit{qap} stands for Question-Answer Pair and \textit{clarificationq} for Clarification Question. The two main clusters align with the two-dimensional plot of the unweighted UMAP coordinates in Figure \ref{fig:umap} }
    \label{fig:clust_no_conf}
\end{figure}

\section{Annotation Interface} \label{interface}

\begin{figure*}[h]
    \centering
    \includegraphics[width=.9\textwidth]{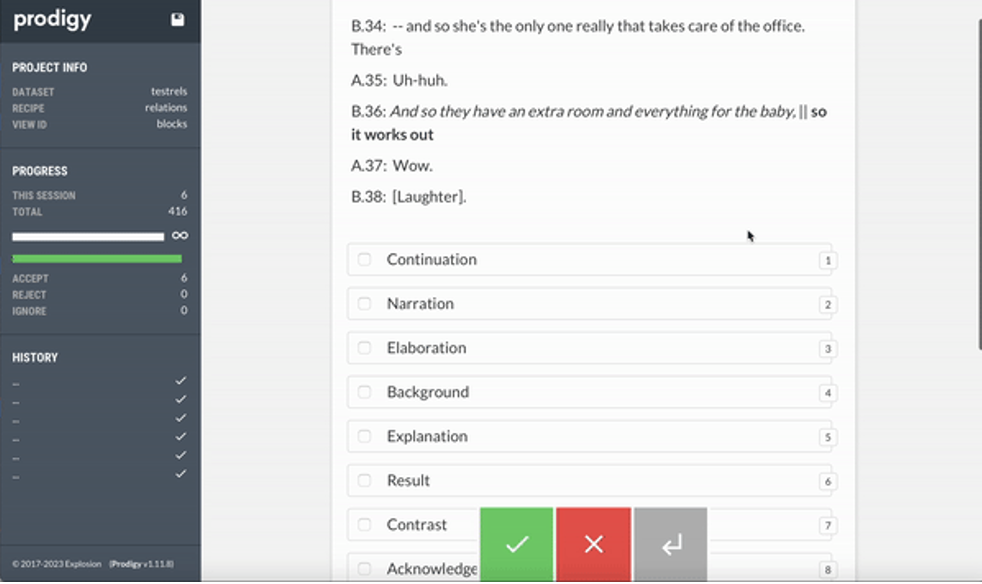}
    \caption{Example annotation task. EDUs to be annotated and discourse relations.}
    \label{fig:my_label}
\end{figure*}

\begin{figure*}[h]
    \centering
    \includegraphics[width=.9\textwidth]{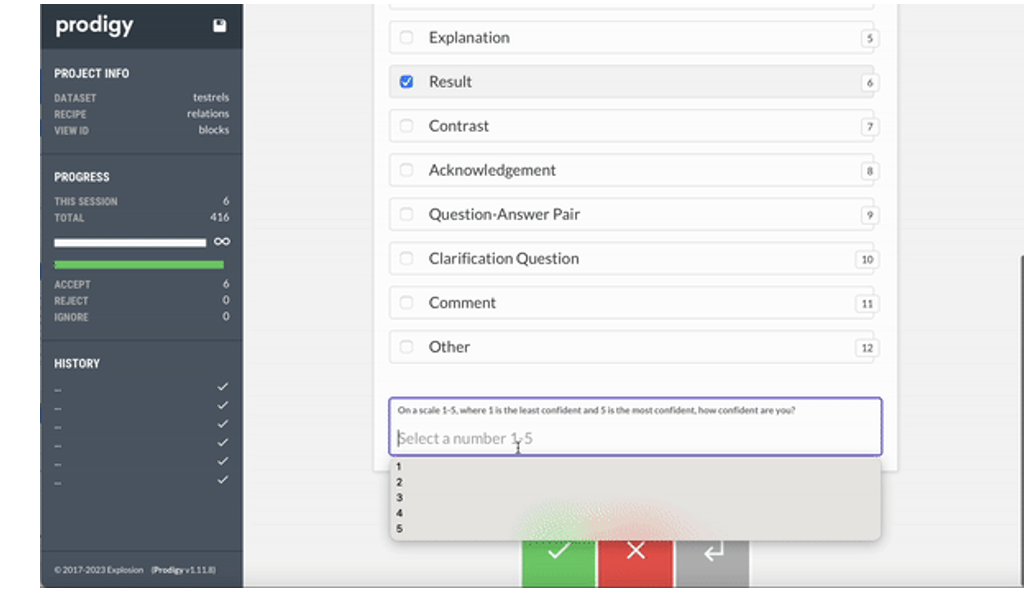}
    \caption{Example annotation task. Discourse relations and confidence score.}
    \label{fig:my_label}
\end{figure*}

\end{document}